\documentclass[12pt]{article}

\usepackage[letterpaper, margin=1in]{geometry}

\usepackage{lmodern}
\usepackage[utf8]{inputenc} % allow utf-8 input
\usepackage[T1]{fontenc}    % use 8-bit T1 fonts
\usepackage{hyperref}       % hyperlinks
\usepackage{url}            % simple URL typesetting
\usepackage{booktabs}       % professional-quality tables
\usepackage{amsfonts}       % blackboard math symbols
\usepackage{nicefrac}       % compact symbols for 1/2, etc.
\usepackage[activate={true,nocompatibility},final,tracking=true,kerning=true,spacing=true,factor=1100,stretch=10,shrink=10]{microtype}      % microtypography
\usepackage{graphicx}
\usepackage[list=true]{subcaption}
\usepackage{lipsum}
\usepackage{amsmath}
\usepackage{mathtools}
\usepackage{amsthm}
\usepackage{amssymb}
\usepackage{footnote}
\usepackage{tablefootnote}
\usepackage[bottom]{footmisc}
\usepackage[backend=bibtex,sorting=nty,sortcites,maxbibnames=99]{biblatex}
\usepackage{algorithm}
\usepackage{algorithmicx}
\usepackage{algpseudocode}
\usepackage{tikz}
\usepackage{enumitem}
\usepackage{threeparttable}
\usepackage{multirow}
\usepackage{longtable}
\usepackage{chngcntr}
\usepackage{appendix}
\usepackage{setspace}
\usepackage[frozencache=true,cachedir=.]{minted}
\usepackage[usestackEOL]{stackengine}
\stackMath
\setstackgap{L}{14pt}

\newtheoremstyle{mytheoremstyle}{5pt}{0pt}{\itshape}{}{\bfseries}{.}{.5em}{} 
\theoremstyle{mytheoremstyle}

\newtheorem{lemma}{Lemma}
\newtheorem{definition}{Definition}
\AtBeginDocument{%
  \setlength{\abovedisplayskip}{4pt}%
  \setlength{\belowdisplayskip}{4pt}%
  \setlength{\abovedisplayshortskip}{0pt}%
  \setlength{\belowdisplayshortskip}{0pt}%
}

\expandafter\def\expandafter\UrlBreaks\expandafter{\UrlBreaks%  save the current one
  \do\a\do\b\do\c\do\d\do\e\do\f\do\g\do\h\do\i\do\j%
  \do\k\do\l\do\m\do\n\do\o\do\p\do\q\do\r\do\s\do\t%
  \do\u\do\v\do\w\do\x\do\y\do\z\do\A\do\B\do\C\do\D%
  \do\E\do\F\do\G\do\H\do\I\do\J\do\K\do\L\do\M\do\N%
  \do\O\do\P\do\Q\do\R\do\S\do\T\do\U\do\V\do\W\do\X%
  \do\Y\do\Z}

\title{Survival regression with accelerated failure time model in XGBoost}

\author{%
  Avinash Barnwal\thanks{
    The author gratefully acknowledges Google for supporting him via Google Summer of Code 2019.}\\
  Stony Brook University\\
  \texttt{avinashbarnwal123@gmail.com}
  \and
  Hyunsu Cho \\
  NVIDIA \\
  \texttt{phcho@nvidia.com}
  \and
  Toby Hocking \\
  Northern Arizona University\\
  \texttt{toby.hocking@nau.edu}
}

\begin{document}

\maketitle

\begin{abstract}
Survival regression is used to estimate the relation between time-to-event and feature variables, and is important in application domains such as medicine, marketing, risk management and sales management. 
Nonlinear tree based machine learning algorithms as implemented in libraries such as XGBoost, scikit-learn, LightGBM, and CatBoost are often more accurate in practice than linear models. However, existing state-of-the-art implementations of tree-based models have offered limited support for survival regression. 
In this work, we implement loss functions for learning accelerated failure time (AFT) models in XGBoost, to increase the support for survival modeling for different kinds of label censoring. 
We demonstrate with real and simulated experiments the effectiveness of AFT in XGBoost with respect to a number of baselines, in two respects: generalization performance and training speed. Furthermore, we take advantage of the support for NVIDIA GPUs in XGBoost to achieve substantial speedup over multi-core CPUs. To our knowledge, our work is the first implementation of AFT that utilizes the processing power of NVIDIA GPUs. Starting from the 1.2.0 release, the XGBoost package natively supports the AFT model. The addition of AFT in XGBoost has had significant impact in the open source  community, and a few statistics packages now utilize the XGBoost AFT model.
\end{abstract}

\onehalfspacing
\section{Introduction}
\label{sec:intro}
Survival analysis is a prominent subfield of statistics, where the goal is to model time duration to a given event (e.g. death). Given the nature of time-to-event data, labels may not be completely observed and only censored labels are given. For data points whose label is censored, the exact label $y$ is not known but only a range $(\underline{y}, \overline{y})$ that contains it. The topic has drawn a large body of research literature in the last few decades. See \cite{survival_survey} for a general survey.

The Cox proportional hazards (Cox-PH) model \cite{cox_ph} is one of the most commonly used models in survival analysis. 
The model estimates the hazard function $h(t)$, which is defined to be the likelihood of the event occurring at time $t$ given that no event has occurred before time $t$. 
The Cox-PH model is of form $h(t, \mathbf{x}) = h_0(t) \exp{(\langle \mathbf{w}, \mathbf{x} \rangle)}$, where the baseline hazard function $h_0(t)$ depends only on time and the features $\mathbf{x}$ have multiplicative effects on $h$. 
Given the parameters $\mathbf{w}$, it is clear which of the normalized features $\mathbf x$ has the largest effect on survival. 
However, it is non-trivial to predict time-to-event $\hat{y}(\mathbf{x})$ in the Cox-PH model. 
We would need to estimate the baseline hazard function $h_0(t)$ using a non-parametric estimator known as Breslow's estimator \cite{breslow1972, allison2010survival}. 
The computation of Breslow's estimator requires access to the full training data and is computationally expensive for big data. 
%For the rest of this paper we assume the big data setting, so in particular we do not include Cox-PH in our experimental comparisons because it does not yield a prediction $\hat{y}$ directly.

The Accelerated Failure Time (AFT) model is another well known method for survival analysis, although perhaps less often used than Cox-PH. 
We choose to explore AFT in this paper for two primary reasons.
First, we would like to not only analyze model parameters (coefficients) but also perform predictive analysis. 
While Cox-PH gives relative importance of features, it does not yield a usable prediction $\hat{y}$ easily \cite{allison2010survival}. 
With the AFT model, we can predict unknown labels using only the fitted parameters and a feature vector. 
Second, the AFT model may provide a better fit when proportional hazard assumption does not hold \cite{aft_vs_coxph}.

Miller \cite{Miller1976} proposed the AFT model for the first time, and later Buckley and James \cite{BuckleyJames1979} refined it to obtain an asymptotically consistent estimator using the least squares approach. Khan and Shaw \cite{khan2016} combined AFT with adaptive and weighted elastic nets to enable variable selection from high-dimensional data. See \cite{aft_survey, aftgee} for overviews on the topic of AFT models.

Tree-based models have shown better performance than linear models in terms of detecting complex and nonlinear patterns in the feature variables. 
The gradient boosting algorithm \cite{Friedman} fits an additive ensemble of decision trees in a stepwise fashion to greedily optimize a general class of loss functions $\ell(y, \hat{y})$. 
Gradient boosting is widely used due to its simplicity and predictive performance. 
The algorithm produces an ensemble of decision trees and exhibits many desirable properties as a statistical model, such as being slow to overfitting and having asymptotic convergence guarantees \cite{BoostingStats, boosting-convergence}. 
Gradient boosting is versatile, as it can optimize a general class of loss function $\ell(y, \hat{y})$ where $y$ represents the true label and $\hat{y}$ the predicted label. 
It has been successfully used in classification \cite{friedman2000}, document ranking \cite{burges2010}, structured prediction \cite{pmlr-v38-chen15b} and other applications. 
Today, there are several scalable, efficient software packages that implement gradient boosting, including XGBoost \cite{XGBoost2016KDD}, LightGBM \cite{LightGBM2017NIPS}, Scikit-Learn \cite{scikit-learn}, and Catboost \cite{CatBoost2018NIPS}.

XGBoost is a fast implementation of gradient boosting that speeds up convergence by using the second-order partial derivative of the loss function. 
XGBoost is able to integrate with a variety of programming environments such as R and Python and integrates with frameworks for distributed computing, such as Dask and Apache Spark. 
Integration of AFT with XGBoost therefore makes survival analysis easier in the big data setting.

There are a few previous implementations of survival analysis in tree based models. 
Schmid and Hothorn \cite{Schmid2008FlexibleBO} used boosting framework for AFT and considers the negative log-likelihood as loss function. 
There are also other tree based survival models such as Random Survival Trees \cite{ishwaran2008}, Cox-Boosting \cite{CoxBoost}, Bagging Survival Trees \cite{BaggSurvTree}, Scikit-Survival \cite{sksurv} and Cox-PH in XGBoost \cite{lundberg2018consistent}. 
Most of the models are limited to right-censored outcomes.
Maximum Margin Interval Trees \cite{Drouin2017} support interval-censored labels.

Survival analysis is broadly useful in a variety of applications, such as survival prediction of cancer patients \cite{Vigan2000}, customer churn \cite{VanDenPoel2004}, credit scoring \cite{Dirick2017}, failure times of mechanical systems \cite{Susto2015, Barabadi2010}. 
However, binary machine learning classifiers have been often used where survival methodology is applicable, due to concerns about predictive accuracy \cite{coxneural}. 
For example, Vaid et al. \cite{Covid19NYCResponse} used XGBoost binary classifiers to predict whether COVID-19 patients will develop complications in a given time frame, achieving substantially better AUC-ROC and AUC-PR than generalized linear models. 
While binary classifiers may provide for a state-of-art predictive accuracy, one loses flexibility that comes from directly modeling time duration to events: one is forced to decide predetermined duration(s) where an event is to occur or not. 
AFT in XGBoost addresses these challenges in the following two ways. 
First, the model is able to capture nonlinear patterns in the data. 
Second, the model readily produces survival time estimates; to compute predictions, we only need the fitted model parameters and a feature vector.

\paragraph{Summary of novel contributions.} We propose a novel adaptation of the AFT model to integrate with XGBoost. 
Our implementation supports all modes of label censoring, including interval-censoring. 
We run experiments with real and simulated data sets to demonstrate the generalization performance of the AFT model in XGBoost. 
Furthermore, we are able to accelerate training by using XGBoost's built-in support for NVIDIA GPUs.

\section{AFT in XGBoost}
The original AFT model takes the following form:
\begin{equation}
\label{original-aft}
\ln{Y} = \langle \mathbf{w}, \mathbf{x} \rangle + \sigma Z
\end{equation}
where $\mathbf{x}$ represents the input features, $\mathbf{w}$ the coefficients, $Y$ the survival time (the output label), and $Z$ a random variable of a known probability distribution. Both $Y$ and $Z$ are random variables. Note that this model is a generalized form of a linear regression model $Y = \langle \mathbf{w}, \mathbf{x} \rangle$. In order to make AFT work with gradient boosting, we revise the model as follows:
\begin{equation}
\label{revised-aft}
\ln{Y} = \mathcal{T}(\mathbf{x}) + \sigma Z
\tag{\ref{original-aft}'}
\end{equation}
where $\mathcal{T}(\mathbf{x})$ represents the output from the decision tree ensemble, given input $\mathbf{x}$.

\subsection{Derivation of AFT loss function}

XGBoost optimizes a twice-differentiable convex loss function $\ell(\cdot, \cdot)$ in its second-order method of gradient boosting \cite{XGBoost2016KDD}. We will now define a suitable loss function $\ell_{\mathrm{AFT}}$ to represent the AFT model. Let $\mathcal{D} = \{(\mathbf{x}_i, y_i)\}_{i=1}^n$ denote the training data, and let $Y_1, \ldots, Y_n$ denote random variables i.i.d. with the distribution for $Y$.
Let $f_Y$ and $F_Y$ denote the probability density function (PDF) and the cumulative distribution function (CDF) for $Y$, respectively.
The likelihood function for $\mathcal{D}$ is the product of probability densities $f_Y$ for individual data points:
\begin{equation}
L(\mathcal{D}) = \mathbb{P}[Y_1 = y_1, \ldots, Y_n = y_n] = \prod_{i=1}^n \mathbb{P}[Y_i = y_i] = \prod_{i=1}^n f_Y(y_i)
\label{likelihood}
\end{equation}
As commonly done in machine learning literature, we maximize log likelihood instead:
\begin{equation}
\ln{L(\mathcal{D})} = \sum_{i=1}^n \ln{\mathbb{P}[Y_i = y_i]} = \sum_{i=1}^n \ln{f_Y(y_i)}
\label{log-likelihood}
\tag{\ref{likelihood}'}
\end{equation}
Since we do not know $y_i$ for some data points, due to label censoring, we revise the likelihood function (\ref{log-likelihood}) to take account of the censored labels:
\begin{align*}
\ln{L(\mathcal{D})} &= \underbrace{\sum \ln{\mathbb{P}[Y_i = y_i]}}_{\text{uncensored label}}
                    + \underbrace{\sum \ln{\mathbb{P}[\underline{y_i} \leq Y_i \leq \overline{y_i}]}}_{\text{censored label with }y_i\in [\underline{y_i}, \overline{y_i}]}\\
                    &= \underbrace{\sum \ln{f_Y(y_i)}}_{\text{uncensored label}}
                    + \underbrace{\sum \ln{(F_Y(\overline{y_i}) - F_Y(\underline{y_i}))}}_{\text{censored label with }y_i\in [\underline{y_i}, \overline{y_i}]}
\end{align*}
where $\underline{y_i}$ and $\overline{y_i}$ are lower and upper bounds for the label $y_i$, respectively. Note that $\overline{y_i}$ may be infinity, to indicate right-censored labels. See Table~\ref{censoring-types} for full list of censoring types. We are now ready to define the loss function $\ell_{\mathrm{AFT}}$.

\begin{table}
\caption{List of label censoring types}
\label{censoring-types}
\centering
\begin{tabular}{ccc}\toprule
Label censoring & Lower bound ($\underline{y_i}$) & Upper bound ($\overline{y_i}$)\\\midrule
Right-censored & Finite non-negative & $+\infty$\\
Left-censored & 0 & Finite non-negative\\
Interval-censored & Finite non-negative & Finite non-negative\\\bottomrule
\end{tabular}
\end{table}

\begin{figure}
    \centering
    \includegraphics[width=\linewidth]{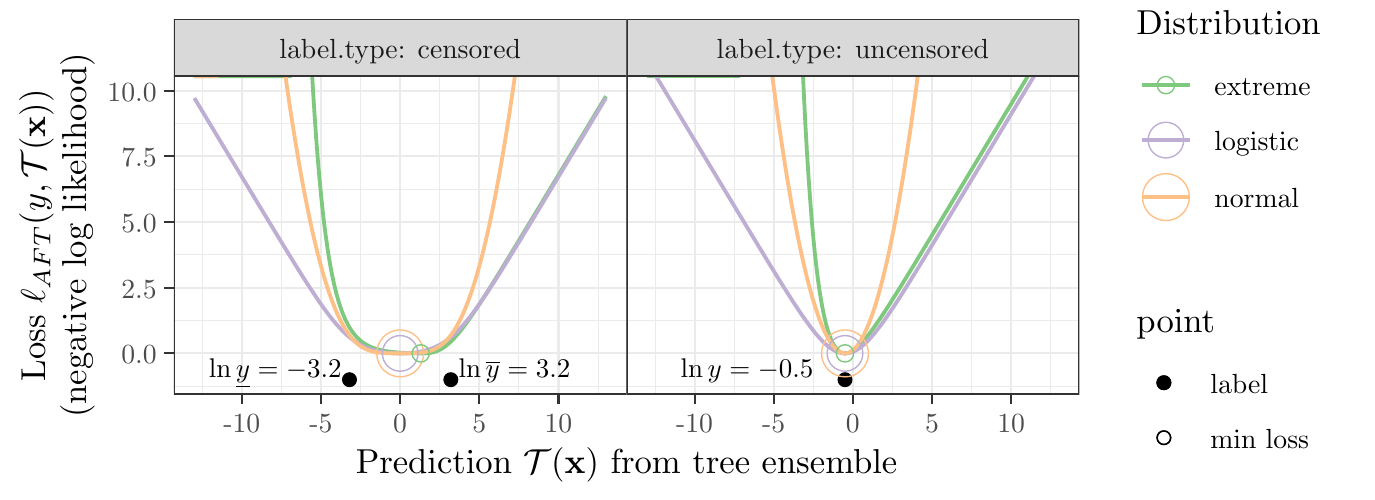}
    \vskip -0.5cm 
    \caption{Geometric interpretation of the Accelerated Failure Time (AFT) loss (colored curves), using three distributions (normal, logistic, extreme) with scale parameter $\sigma=1$.
    Log survival times are shown using solid black dots, and loss function minima are shown using open colored circles. Note that the prediction $\mathcal{T}(\mathbf{x})$ from the tree ensemble model is in the same scale as the log survival time $\ln{y}$.
    \textbf{Left:} for censored data the loss function is defined as the negative log of the difference of cumulative distribution function values. The example shown has finite upper and lower limits, for which the minimum of the logistic/normal loss occurs at the midpoint between the two limits, whereas it occurs at a greater value for the extreme distribution. \textbf{Right:} for uncensored data the loss function is defined as the negative log of the density function, so the normal loss is the usual square loss with symmetric quadratic tails. The logistic loss has symmetric linear tails, whereas the asymmetric extreme loss has a linear upper tail and an exponential lower tail.}
    \label{fig:loss-new}
\end{figure}

\begin{definition}[Loss function for AFT survival regression]
\begin{equation}
\ell_{\mathrm{AFT}}(y,\mathcal{T}(\mathbf{x})) = 
\begin{cases}
-\ln{f_Y(y)} & \text{if $y$ is not censored} \\
-\ln{(F_Y(\overline{y}) - F_Y(\underline{y}))} & \text{if $y$ is censored with $y\in[\underline{y}, \overline{y}]$}
\end{cases}
\label{aft-loss}
\end{equation}
\end{definition}
Under this definition, the sum of losses $\sum_{i=1}^n \ell(y_i, \mathcal{T}(\mathbf{x}_i))$ over the training data is identical to $-\ln{L(\mathcal{D})}$. Since we only know distribution of $Z$ (not of $Y$), we use the following lemma:
\begin{lemma}[(1.27) of \cite{Bishop2006}]
Let $Y$ and $Z$ be random variables. If $Y = g(Z)$ with a monotone increasing function $g(\cdot)$ that is suitably smooth, the PDF and CDF of $Y$ are expressed in terms of the PDF and CDF of $Z$ as follows:
\begin{align}
f_Y(y) &= f_Z(g^{-1}(y))\cdot \frac{d}{dy}g^{-1}(y) &
F_Y(y) &= F_Z(g^{-1}(y))
\end{align}
\label{variable-change}
\end{lemma}
We apply Lemma~\ref{variable-change} to (\ref{aft-loss}) with $g(Z) = \exp{(\mathcal{T}(\mathbf{x}) + \sigma Z)}$ to get the following formula for $\ell_{\mathrm{AFT}}$:
\begin{definition}[Loss function for AFT survival regression, in terms of known PDF and CDF]
\begin{equation}
\ell_{\mathrm{AFT}}(y,\mathcal{T}(\mathbf{x})) = 
\begin{cases}
-\ln{\left[f_Z(s(y)) \cdot \dfrac{1}{\sigma y}\right]} & \text{if $y$ is not censored}  \\[3ex]
-\ln{\left[F_Z(s(\overline{y})) - F_Z(s(\underline{y}))\right]} & \text{if $y$ is censored with $y\in[\underline{y}, \overline{y}]$}
\end{cases}
\label{aft-loss-concrete}
\tag{\ref{aft-loss}'}
\end{equation}
where $f_Z$ and $F_Z$ are given by Table~\ref{survival-distributions} and $s(y) = s(y, \mathcal{T}(\mathbf{x})) = (\ln{y} - \mathcal{T}(\mathbf{x}))/\sigma$ is a link function.
\label{aft-loss-concrete-definition}
\end{definition}
See Figure~\ref{fig:loss-new} for a geometric representation. Since the prediction $\mathcal{T}(\mathbf{x})$ from the tree ensemble model approximates the log survival time $\ln{y}$, we use the link function $s(y, \mathcal{T}(\mathbf{x})) = (\ln{y} - \mathcal{T}(\mathbf{x})) / \sigma$ in Definition~\ref{aft-loss-concrete-definition} as a convenient measure for the distance between the prediction and the log survival time. Although $s$ is a function of two variables, we will use $s(y)$ as a shorthand for $s(y, \mathcal{T}(\mathbf{x}))$ to save space.

\subsection{Gradient and hessian of the AFT loss}
The gradient boosting algorithm in XGBoost uses the gradient and hessian of the loss function, which are first and second partial derivatives of $\ell(y, \mathcal{T}(\mathbf{x}))$ with respect to the second input $\mathcal{T}(\mathbf{x})$. To express partial derivatives in a concise manner, define a single-letter variable $u = \mathcal{T}(\mathbf{x})$ as an alias for the output from the tree ensemble model. The gradient and hessian of the AFT loss function are as follows\footnote{For left- and right-censored labels, let $f_Z(-\infty) = f_Z(\infty) = 0$ and $F_Z(-\infty) = 0, F_Z(+\infty) = 1$.}:
\begin{definition}[Gradient and hessian of AFT loss]\label{def:grad-hess-aft}
\begin{align}
\left.\frac{\partial\ell_{\mathrm{AFT}}}{\partial u}\right|_{y,u}
&= 
\begin{cases}
\dfrac{f'_Z(s(y))}{\sigma f_Z(s(y))} & \text{if $y$ is not censored}  \\[3ex]
\dfrac{f_Z(s(\overline{y})) - f_Z(s(\underline{y}))}{\sigma [ F_Z(s(\overline{y})) - F_Z(s(\underline{y})) ] } & \text{if $y$ is censored with $y\in[\underline{y}, \overline{y}]$}
\end{cases}\label{def:grad-aft}\\[4ex]
\left.\frac{\partial^2 \ell_{\mathrm{AFT}}}{\partial u^2}\right|_{y,u}
&= 
\begin{cases}
-\dfrac{f_Z(s(y)) f_Z''(s(y)) - f_Z'(s(y))^2}{\sigma^2 f_Z(s(y))^2} & \text{if $y$ is not censored}  \\[3ex]
\dfrac{\Longstack[l]{ -[ F_Z(s(\overline{y})) - F_Z(s(\underline{y})) ][ f_Z'(s(\overline{y})) - f_Z'(s(\underline{y})) ] \\ \phantom{-} + [ f_Z(s(\overline{y})) - f_Z(s(\underline{y})) ]^2 } }{\sigma^2 [ F_Z(s(\overline{y})) - F_Z(s(\underline{y})) ]^2 } & \text{if $y$ is censored}
\end{cases}\label{def:hess-aft}
\end{align}
where $f_Z'$ and $f_Z''$ are the first and second derivatives of the PDF $f_Z$, respectively, and $s(y) = (\ln{y} - u)/\sigma$ is defined the same way as in Definition~\ref{aft-loss-concrete-definition}. See Table~\ref{survival-distributions} to look up $f_Z'$ and $f_Z''$ for the three known distributions. 
\end{definition}
\begin{proof}
The first- and second-order partial derivatives of $\ell_{\mathrm{AFT}}$ may be derived using basic rules of Calculus. Consult Appendix~\ref{appendix:loss_gradient_full_proof} for the full proof.
\end{proof}

\begin{table}[t]
\caption{Probability distributions for $Z$}
\label{survival-distributions}
\centering
\begin{threeparttable}
\begin{tabular}{ccccc}\toprule
Distribution & PDF ($f_Z(z)$) & CDF ($F_Z(z)$) & $f_Z'(z)$ & $f_Z''(z)$\\\midrule
Normal & $\dfrac{\exp{(-z^2/2)}}{\sqrt{2\pi}}$ & $\dfrac{1}{2}\left(1 + \mathrm{erf}\left(\dfrac{z}{\sqrt{2}}\right)\right)$ & $-zf_Z(z)$ & $(z^2 - 1)f_Z(z)$ \\[3ex]
Logistic & $\dfrac{e^z}{(1+e^z)^2}$ & $\dfrac{e^z}{1+e^z}$ & $\dfrac{f_Z(z)(1 - e^z)}{1 + e^z}$ & $\dfrac{f_Z(z) (e^{2z} - 4e^z + 1)}{(1 + e^z)^2}$ \\[3ex]
Extreme\tnote{1}& $e^z e^{-\exp{z}}$ & $1 - e^{-\exp{z}}$ & $(1 - e^z)f_Z(z)$ & $(e^{2z} - 3e^z + 1)f_Z(z)$ \\\bottomrule
\end{tabular}
\begin{tablenotes}
\footnotesize
\item[1] Also known as the Gumbel (minimum) distribution. See \cite{survival-package}.
\end{tablenotes}
\end{threeparttable}
\end{table}

\subsection{Regularization for the AFT loss, to avoid numerical instabilities}\label{sec:regularization}
The equations (\ref{aft-loss-concrete}), (\ref{def:grad-aft}), and (\ref{def:hess-aft}) may suffer from numerical instabilities when the prediction $u = \mathcal{T}(\mathbf{x})$ from the tree ensemble model is far away from the true log survival time $\ln{y} \in [\ln{\underline{y}}, \ln{\overline{y}}]$. There are three causes for the numerical instabilities:
\begin{itemize}
    \item Zero passed to the logarithm function $\ln{(\cdot)}$: the difference term $F_Z(s(\overline{y})) - F_Z(s(\underline{y}))$ in (\ref{aft-loss-concrete}) becomes nearly zero when the prediction $u = \mathcal{T}(\mathbf{x})$ is far away from the true log survival time $\ln{y} \in [\ln{\underline{y}}, \ln{\overline{y}}]$. Passing zero to the logarithm results in a NAN (Not-a-Number).
    \item Zero denominator: the denominators in (\ref{def:grad-aft}), and (\ref{def:hess-aft}) for censored data contain the difference term $F_Z(s(\overline{y})) - F_Z(s(\underline{y}))$, which can be nearly zero for the same reason as above. Zero denominator results in a NAN (Not-a-Number).
    \item Large number passed to the exponential function $\exp{(\cdot)}$: the PDF and CDF of the logistic and extreme distributions contain the exponential function. Since 64-bit floating-point variables in a C++ program hold up to $10^{308}$, the exponential function yields in an INF (infinity) even for moderately large inputs.
\end{itemize}
Refer to the IEEE 754 standard \cite{IEEE754} to learn more about the special representations of floating-point values, such as INFs and NANs. In order to eliminate INFs and NANs, we apply regularization in two places.

\subsubsection{Regularization for the loss function (\ref{aft-loss-concrete})}
We replace the difference term $F_Z(s(\overline{y})) - F_Z(s(\underline{y}))$ with $\epsilon = 10^{-12}$ whenever the difference term is smaller than $\epsilon$.

\subsubsection{Regularization for the gradient (\ref{def:grad-aft}) and the hessian (\ref{def:hess-aft})}
We explicitly define values of the gradient and hessian at the limit $u \to \pm\infty$. Whenever a numerical instability is detected due to $u$ being far away from the true log survival time, we set the gradient and hessian values according to Table~\ref{appendix:table:derivatives_at_extreme}. We clip all gradients to $\pm 15$, as large gradients cause numerical difficulties in XGBoost. In addition, we always assign $10^{-16}$ hessian value to all data points, because data points with zero hessian are ignored by XGBoost. Regularization forces XGBoost to consider every data point to a certain extent.

\begin{table}[ht]
\caption{Specification of the gradient and hessian values ($\partial \ell / \partial u$, $\partial^2 \ell / \partial u^2$) as $u \to \pm\infty$.}
\label{appendix:table:derivatives_at_extreme}
\centering
\small
\begin{tabular}{rrrrrr}\toprule
\multirow{2}{*}[-3pt]{Distribution for $Z$} & Label type	&	\multicolumn{2}{c}{Uncensored}	&	\multicolumn{2}{c}{Right-censored}\\\cmidrule{2-6}
& & $u\to-\infty$ &	$u\to+\infty$ &	$u\to-\infty$ &	$u\to+\infty$ \\\midrule
\multirow{2}{*}[0pt]{Normal} &	$\partial \ell / \partial u$ &	$-15$ &	15&	$-15$&	0\\
       &	$\partial^2 \ell / \partial u^2$ &	$1/\sigma^2$&	$1/\sigma^2$&	$1/\sigma^2$&	$10^{-16}$\\
\multirow{2}{*}[0pt]{Logistic} &	$\partial \ell / \partial u$ &	$-1/\sigma$&	$1/\sigma$&	$-1/\sigma$&	0\\
&	$\partial^2 \ell / \partial u^2$&	$10^{-16}$&	$10^{-16}$&	$10^{-16}$&	$10^{-16}$\\
\multirow{2}{*}[0pt]{Extreme} &	$\partial \ell / \partial u$ &	$-15$&	$1/\sigma$&	$-15$&	0\\
 &	$\partial^2 \ell / \partial u^2$ &	15&	$10^{-16}$&	15&	$10^{-16}$\\\midrule[\heavyrulewidth]
%% sec row
\multirow{2}{*}[-3pt]{Distribution for $Z$} & Label type	&	\multicolumn{2}{c}{Left-censored}	&	\multicolumn{2}{c}{Interval-censored}\\\cmidrule{2-6}
& &	$u\to-\infty$ &	$u\to+\infty$ &	$u\to-\infty$ &	$u\to+\infty$\\\midrule
\multirow{2}{*}[0pt]{Normal} &	$\partial \ell / \partial u$ &0&	15&	$-15$&	15\\
       &	$\partial^2 \ell / \partial u^2$ &$10^{-16}$&	$1/\sigma^2$&	$1/\sigma^2$&	$1/\sigma^2$\\
\multirow{2}{*}[0pt]{Logistic} &	$\partial \ell / \partial u$ &0&	$1/\sigma$&	$-1/\sigma$&	$1/\sigma$\\
&	$\partial^2 \ell / \partial u^2$&	$10^{-16}$&	$10^{-16}$&	$10^{-16}$&	$10^{-16}$\\
\multirow{2}{*}[0pt]{Extreme} &	$\partial \ell / \partial u$ &0&	$1/\sigma$&	$-15$&	$1/\sigma$\\
 &	$\partial^2 \ell / \partial u^2$ &	$10^{-16}$&	$10^{-16}$&	15&	$10^{-16}$\\\bottomrule
\end{tabular}
\end{table}

\section{Experiments}

\subsection{Effectiveness of AFT for interval-censored data}\label{sec:interval-censored}
We measure the effectiveness of the XGBoost AFT model for interval-censored data. We define the accuracy metric for data with interval-censored labels as follows:
\begin{equation}
    \mathrm{Accuracy}(\mathcal{D}) = \frac{\left|\{i : \mathcal{T}(\mathbf{x}_i) \in [\ln{\underline{y_i}}, \ln{\overline{y_i}}]\}\right|}{\left|\mathcal{D}\right|},
\end{equation}
i.e. the fraction of data points for which the prediction from the tree ensemble model falls between the lower and upper bounds for the true log survival time.

The XGBoost AFT model is compared to three baselines:
\begin{description}
\item[survreg] Un-regularized linear model with AFT loss functions \cite{survival-package} on principal components, with the number of components selected using cross-validation.
\item[penaltyLearning] L1-regularized linear model with squared hinge loss \cite{Rigaill2013}, with the degree of L1 regularization selected by cross-validation.
\item[MMIT] Max Margin Interval Trees \cite{Drouin2017}, which generalizes the well-known Classification and Regression Tree (CART) algorithm of \cite{Breiman1984} to censored outputs. The tree depth is selected using cross-validation.
\end{description}

We perform nested cross-validation to estimate the generalization performance of the model as well as the hyperparameter search procedure. We use 5-fold internal cross-validation to evaluate multiple hyperparameter combinations. Each combination is judged using the mean validation accuracy over the 5 folds. (The mean validation accuracy is also used to determine the number of boosting rounds.) We then perform 4-fold external cross-validation to quantify the generalization performance of the training procedure, as follows: we fit a new model using the best hyperparameter combinations, using all data points except the held-out test set. The test accuracy is evaluated with the held-out test set. For hyperparameter search, we run 100 trials in the random search; see Section~\ref{sec:sensitivity} for details.

Experiments in Sections~\ref{sec:chipseq} and \ref{sec:simulated} were conducted using a workstation with one Intel Core i7-7800X CPU (3.50 GHz, 6 cores) and two DDR4 RAMs (16 GB each, 2133 MHz), running Ubuntu 18.04 LTS.

\subsubsection{Interval-censored data from supervised changepoint detection problems}\label{sec:chipseq}
To test our algorithm in real data sets with interval-censored outputs, we consider a benchmark of supervised peak detection problems in genomic ChIP-seq data \cite{Rigaill2013, Hocking2017}. 
Each of the ten data sets in Table~\ref{dataset-description} corresponds to a set of high-throughput DNA sequencing experiments. 
Expert biologists manually labeled each data set to indicate locations with and without peaks; 
these labels are used to compute the peak detection error rate.
The goal of the ChIP-seq peak detection is to automatically locate peaks given a new DNA sequence, such that peak detection error rate is minimal.
Each data set is named like \verb+H3K4me3_PGP_immune+:
\begin{itemize}
\item The first field (\verb+H3K4me3+) identifies the assay type. Consult the McGill Epigenomics Mapping Centre website\footnote{\url{https://epigenomesportal.ca/edcc/index.html}} for the full list of assay types and their meaning.
\item The second field (\verb+PGP+) is the initials of the expert biologist who provided the labels.
\item The last field (\verb+immune+) identifies a sample set. Consult \cite{Hocking2017} for more information.
\end{itemize}
%In these benchmarks, there are sample-specific feature vectors $\mathbf{x}$ as well as outputs $y$ which are always censored (either left, right, or interval censored; there are no un-censored outputs in these data sets).
A univariate signal is computed for each sample by computing coverage frequency of aligned DNA sequence reads at each positions of the reference genome sequence. 
To detect peaks in the univariate signal, we use an changepoint detection algorithm with learned penalty functions \cite{Rigaill2013}.
Briefly, the penalty parameter $\lambda$ of the changepoint detection algorithm controls the number of distinct segments/peaks detected. 
It is possible to compute a range of optimal penalty values $[\underline{\lambda_{*}}, \overline{\lambda_{*}}]$ that are optimal in terms of the expert-provided labels; that is, setting $\lambda$ to any value in $[\underline{\lambda_{*}}, \overline{\lambda_{*}}]$ will minimize the label error rate.
We cast the peak detection problem into an interval-censored regression problem as follows. 
The univariate signal is used to compute a feature vector $\mathbf{x}\in\mathbb R^{36}$ that stores various summary statistics, such as percentiles, of the signal. 
The range of optimal penalty values $[\underline{\lambda_{*}}, \overline{\lambda_{*}}]$ is taken to be an interval-censored label.

\begin{table}
\caption{Dimensions of ChIP-seq data sets and descriptive statistics. The log lambda values given below are averages over the data set.}
\label{dataset-description}
\centering
\small
\begin{tabular}{rlccrr}\toprule
& Data set & Rows & Columns & \footnotesize{\verb+min.log.lambda+} & \footnotesize{\verb+max.log.lambda+} \\\midrule
(1) & \footnotesize{ATAC\_JV\_adipose} & 465 & 36 & 8.581 & 10.470 \\
(2) & \footnotesize{CTCF\_TDH\_ENCODE} & 182 & 36 & 10.246 & 12.643 \\
(3) & \footnotesize{H3K27ac-H3K4me3\_TDHAM\_BP} & 2008 & 36 & 7.674 & 9.641 \\
(4) & \footnotesize{H3K27ac\_TDH\_some} & 95 & 36 & 9.318 & 10.365 \\
(5) & \footnotesize{H3K36me3\_AM\_immune} & 420 & 36 & 8.955 & 12.723 \\
(6) & \footnotesize{H3K27me3\_RL\_cancer} & 171 & 36 & 14.332 & 16.192 \\
(7) & \footnotesize{H3K27me3\_TDH\_some} & 43 & 36 & 10.617 & 11.389 \\
(8) & \footnotesize{H3K36me3\_TDH\_ENCODE} & 78 & 36 & 8.147 & 9.634 \\
(9) & \footnotesize{H3K36me3\_TDH\_immune} & 84 & 36 & 10.939 & 13.003 \\
(10) & \footnotesize{H3K36me3\_TDH\_other} &  40 & 36 & 9.742 & 12.389 \\\bottomrule
\end{tabular}
\end{table}

We pre-process the data as follows: we apply the exponential function $\exp(\cdot)$ to the output labels \texttt{min.log.lambda} and \texttt{max.log.lambda} to obtain the non-negative interval-censored labels \texttt{min.lambda} and \texttt{max.lambda}. Then we remove all feature columns that either 1) had at least one missing value or 2) had zero variance.

Figure~\ref{chipseq-results} shows the generalization performance (test accuracy) and run time of XGBoost and the baseline packages. XGBoost exhibits competitive generalization performance on par with SurvReg, penaltyLearning, and MMIT. In addition, XGBoost is fast: its run time approaches that of SurvReg and penaltyLearning and significantly smaller than that of MMIT. Considering that SurvReg and penaltyLearning are linear models and MMIT nonlinear, the run-time performance speaks to the efficiency of XGBoost.

\subsubsection{Synthetic interval-censored data generated from known distributions}\label{sec:simulated}
Drouin \cite{Drouin2017} generated synthetic interval-censored data based on three kind of features: sine, absolute and linear. It has a mix of nonlinear and linear features having 200 samples and 20 features in each data set. 
We extend it with three more data sets having more complex nonlinear features. 
We use a random number generator to generate interval-censored data as follows.

First, generate the feature vectors $\mathbf{x}\in \mathbb{R}^{20}$ by sampling from the uniform distribution $U([0, 10])$. Second, draw 10 values randomly from the normal distribution $\mathcal{N}(f(\mathbf{x}), 0.3)$ where the mean is determined with a function $f : \mathbb{R}^{20} \to \mathbb{R}$ that maps the feature vector $\mathbf{x}$ to a real value. Third, out of the 10 values, choose the smallest as the lower bound $\underline{y}$ and the largest as the upper bound $\overline{y}$. Lastly, add a small noise to both the interval bounds by sampling a value from $\mathcal{N}(0, 0.2)$ and adding it to $\underline{y}$ and $\overline{y}$.

Each generated data set was named after the choice for $f$:

\begin{itemize}
    \item \verb+simulated.sin+: $f(\mathbf{x}) = \sin(x_1)$
    \item \verb+simulated.abs+: $f(\mathbf{x}) = \vert x_1 - 5 \vert$
    \item \verb+simulated.linear+: $f(\mathbf{x}) = x_1 / 5$
    \item \verb+simulated.model.1+: $f(\mathbf{x}) = x_1 x_2+x_3^2-x_4 x_7+x_8 x_{10}-x_6^2$
    \item \verb+simulated.model.2+: $f(\mathbf{x}) = -\sin(2x_1)+x_2^2+x_3-\exp(-x_4)$
    \item \verb+simulated.model.3+: $f(\mathbf{x}) = x_1+3x_3^2-2 \exp(-x_5)$
\end{itemize}

\begin{figure}
\begin{subfigure}{\textwidth}
\includegraphics[width=\linewidth]{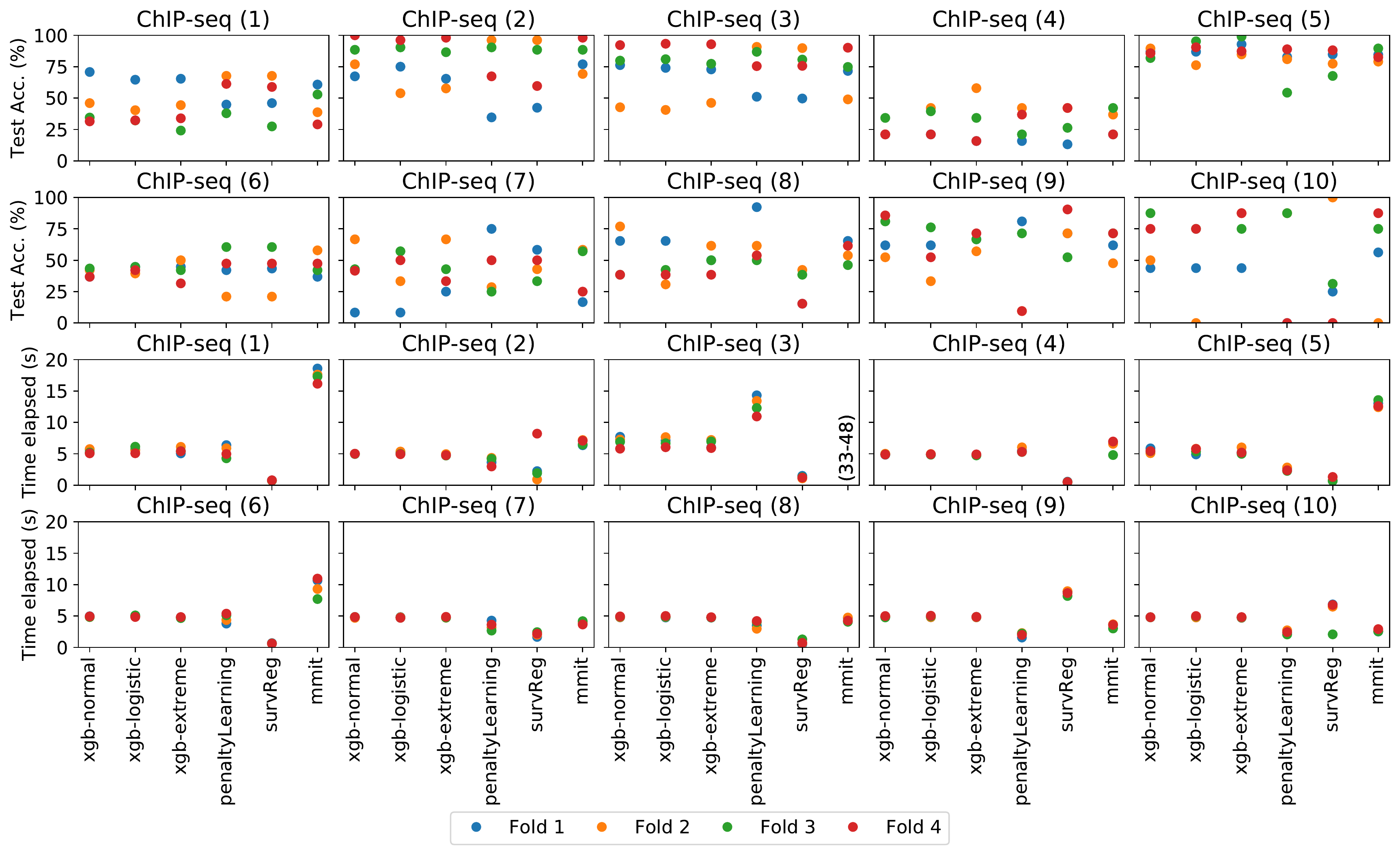}
\caption{ChIP-seq data from Table~\ref{dataset-description}.}
\label{chipseq-results}
\end{subfigure}
\begin{subfigure}{\textwidth}
\includegraphics[width=\linewidth]{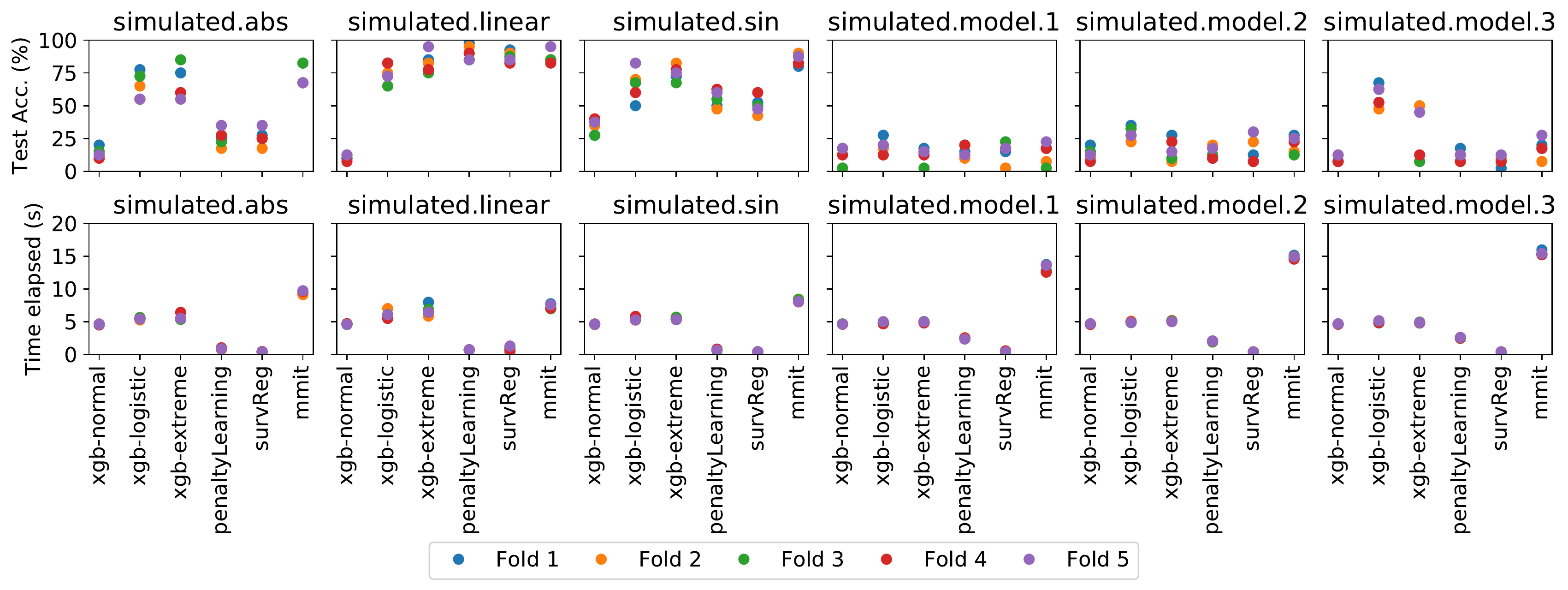}
\caption{Simulated data.}
\label{simulated-results}
\end{subfigure}
\caption{Experimental results from Section~\ref{sec:interval-censored}: test accuracy and run time}
\end{figure}

We compare the performance of penaltyLearning, survReg, MMIT and XGBoost on test data of size 100. As in Section~\ref{sec:chipseq}, we perform nested cross-validation to estimate the generalization performance of the model as well as the hyperparameter search procedure; this time, we use 5 folds for both the outer and inner cross-validation. We reproduce the behavior in \cite{Drouin2017}, where nonlinear models like mmit better capture nonlinear patterns in simulated data than linear models do. In Figure~\ref{simulated-results}, both XGBoost and mmit exhibit superior generalization performance (test accuracy) compared to the linear models, SurvReg and penaltyLearning. The additional run-time incurred by the nonlinear models is compensated by higher test accuracy. The difference between XGBoost and mmit is more pronounced when we look at the three simulated data we added apart from those from \cite{Drouin2017}. XGBoost runs faster than mmit, up to 3x, and shows higher test accuracy. In particular, for \verb+simulated.model.3+, XGBoost achieves 58.5\% mean test accuracy, whereas mmit achieves 18\%.

\subsection{Effectiveness of AFT on right-censored data}\label{sec:right-censored}
We measure the effectiveness of the XGBoost AFT model for right-censored data using Uno's C-statistics \cite{uno2011c}, which is a modified form of Harell's Concorance Index \cite{c-index}. Uno's C-statistics is an unbiased nonparametric estimator for the following ranking metric:
\begin{equation}
C = \mathbb{P}[\mathcal{T}(\mathbf{x}_i) < \mathcal{T}(\mathbf{x}_j) | y_i < y_j, y_i < \tau]
\end{equation}
The $\tau$ constant is chosen judiciously in order to truncate outlier labels when estimating $C$. In this experiment, we set $\tau$ to the 80th percentile of the observed survival time. We use the implementation of Uno's C-statistics from the Scikit-Survival package \cite{sksurv}.

The XGBoost AFT model is compared to two baselines:
\begin{description}
\item[XGBoost-Cox] Cox-PH model from the XGBoost package \cite{lundberg2018consistent}, enabled by setting configuration \verb+objective='survival:cox'+.
\item[Scikit-Survival] Cox-PH linear model from the Scikit-Survival package \cite{sksurv}
\end{description}
As in Section~\ref{sec:interval-censored}, we use nested cross-validation to assess the generalization performance of the model as well as the hyperparameter search procedure. For hyperparameter search, we run 100 trials in the random search; see Section~\ref{sec:sensitivity} for details. 

\subsubsection{Synthetic data with a mix of uncensored and right-censored labels}\label{sec:coxgen}
\begin{figure}[t]
\centering
\includegraphics[width=0.85\linewidth]{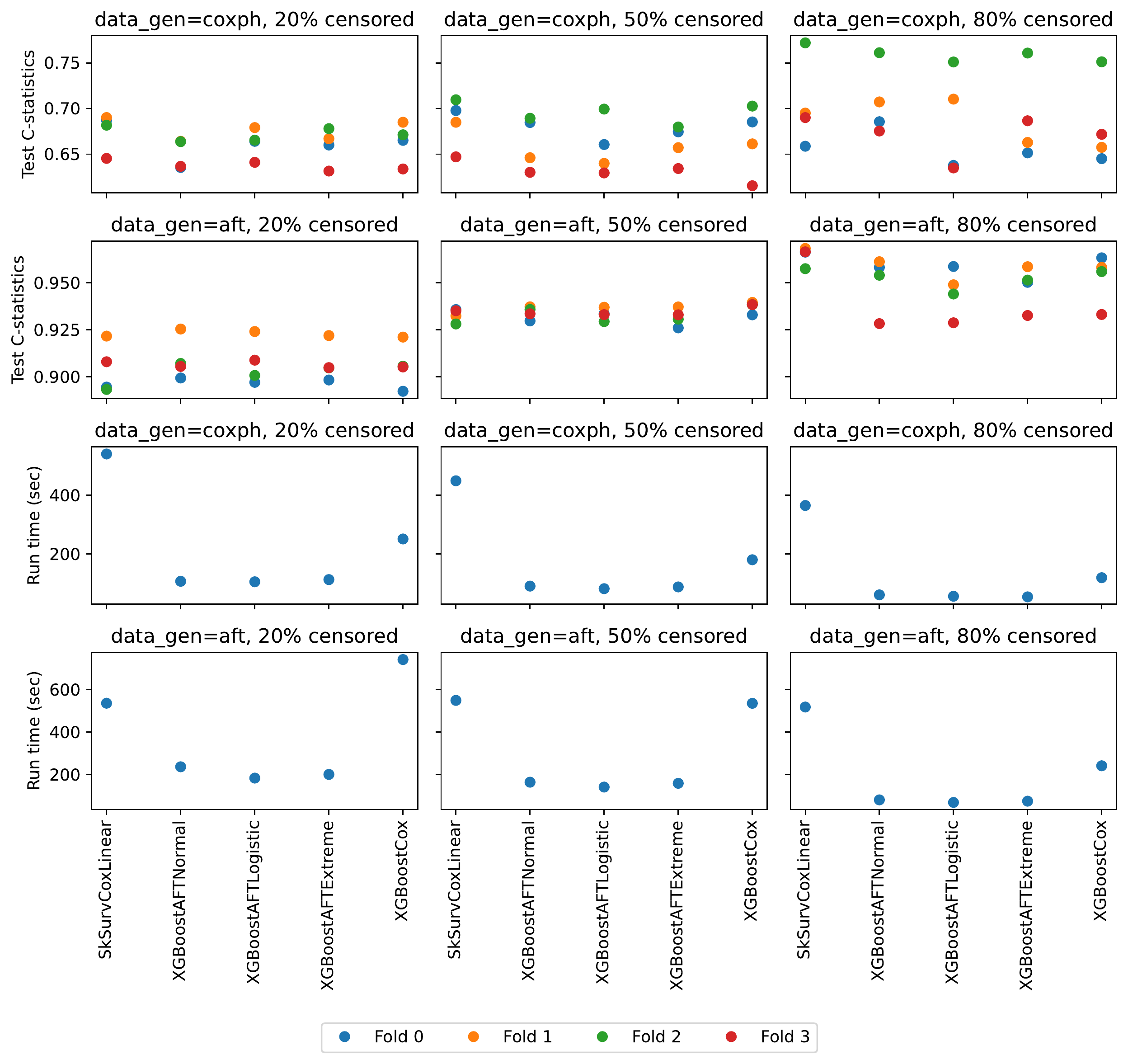}
\caption{Experimental results from Section~\ref{sec:right-censored}: test accuracy and run time}
\label{fig:right-censored-results}
\end{figure}

Using a random number generator, we generate synthetic data consisting a mix of uncensored and right-censored labels, as follows\footnote{This method is adapted from a tutorial on the Scikit-Survival package's website \cite{sksurv}.}:

\begin{enumerate}
    \item Draw a feature vector $\mathbf{x} \in \mathbb{R}^{20}$ from the uniform distribution $U([0, 1])$.
    \item Draw the ``risk score'' $r \in \mathbb{R}$ from the normal distribution $\mathcal{N}(f(\mathbf{x}), 0.3)$ where $f(\mathbf{x}) = x_1 + 3x_3^2 - 2 \exp{(-x_5)}$ is a nonlinear map.
    \item Draw $u$ from the uniform distribution $U([0, 1])$.
    \item Compute the ground-truth survival time $y = -\ln{u} / (h_0 h^r)$, where $h_0 = 0.1$ is the baseline hazard and $h = 2.0$ is the hazard ratio.
    \item Simulate the effect of random censoring by drawing cutoff value $c$ from the uniform distribution $U([0, C])$, where $C$ is suitably chosen to induce censoring for a set fraction of data points. If $y \geq c$, the label is right-censored and we set the label range $[\underline{y}, \overline{y}] = [c, +\infty)$. If $y < c$, the label is not censored and we set $[\underline{y}, \overline{y}] = [y, y]$. 
\end{enumerate}
Repeat the steps to generate 1000 data points. The experiment result with this method of data generation is shown with label \verb+data_gen=coxph+ in Figure~\ref{fig:right-censored-results}. When 20\% the labels are (right-)censored, the Cox-PH model from Scikit-Survival produces slightly higher C-statistics metric than XGBoost models. On the other hand, with greater amount of censoring (50\%, 80\%), the test C-statistics for  XGBoost-AFT and XGBoost-CoxPH are similar to the test C-statistics for Scikit-Survival's Cox-PH.

We now generate data with a mix of uncensored and right-censored labels using a different method.
\begin{enumerate}
    \item Draw a feature vector $\mathbf{x} \in \mathbb{R}^{20}$ from the uniform distribution $U([0, 1])$.
    \item Draw the ``risk score'' $r \in \mathbb{R}$ from the normal distribution $\mathcal{N}(f(\mathbf{x}), 0.3)$ where $f(\mathbf{x}) = x_1 + 3x_3^2 - 2 \exp{(-x_5)}$ is a nonlinear map.
    \item Compute the ground-truth survival time $y = \exp{(-r)}$.
    \item Simulate the effect of random censoring by drawing cutoff value $c$ from the uniform distribution $U([0, C])$. This step is analogous to Step 5 of the previous recipe.
\end{enumerate}
Repeat the steps to generate 1000 data points. The experiment result with this method of data generation is shown with label \verb+data_gen=aft+ in Figure~\ref{fig:right-censored-results}. When 20\% of the labels are censored, XGBoost-AFT with the normal distribution produces slightly higher C-statistics metric than all other models. On the other hand, when 50\% of the labels were censored, there is no clear winner; XGBoost-AFT and XGBoost-CoxPH produce similar test C-statistics as Scikit-Survival's Cox-PH. With 80\% censoring, Scikit-Survival's Cox-PH produces the highest test C-statistics, and XGBoost-AFT and XGBoost-Cox are slightly behind.
In all settings, XGBoost-AFT runs 2-3$\times$ faster than XGBoost-Cox-PH or Scikit-Survival's Cox-PH.

\subsection{Effect of hyperparameters on model performance}\label{sec:sensitivity}
To measure how sensitive the generalization performance is to the choice of hyperparameters, we try an array of approaches for selecting hyperparameters. There are 6 relevant hyperparameters: \verb+learning_rate+, \verb+max_depth+, \verb+min_child_weight+, \verb+reg_alpha+, \verb+reg_lambda+, and \verb+aft_loss_distribution_scale+\footnote{$\sigma$ in (\ref{original-aft})}. The following methods are considered:
\begin{description}
    \item[Grid search] We select one or two hyperparameters out of the six and exhaustively enumerate all combinations using the grid in Table~\ref{hyperparameter-space}. If a hyperparameter is not chosen for the grid search, we assign a default value as follows:
    \verb+learning_rate+ = 0.1, \verb+max_depth+ = 6, \verb+min_child_weight+ = 1.0, \verb+reg_alpha+ = 0.001, \verb+reg_lambda+ = 1.0, \verb+aft_loss_distribution_scale+ = 1.0.
    \item[Random search] We use Optuna~\cite{Optuna} to randomly sample hyperparameter combinations from the search space (Table~\ref{hyperparameter-space}). All six hyperparameters are sampled. Each search is run for 100 or 1000 combinations.
    \item[Baseline (do nothing)] Choose default values for all hyperparameters and perform no search.
\end{description}

\begin{table}
\caption{Search space for hyperparameters}
\label{hyperparameter-space}
\centering
\begin{tabular}{lr}\toprule
Hyperparameter & Search grid\\\midrule
\verb+learning_rate+ & 0.001, 0.01, 0.1, 1.0\\
\verb+max_depth+ & 2, 3, 4, 5, 6, 7, 8, 9, 10\\
\verb+min_child_weight+ & 0.001, 0.1, 1.0, 10.0, 100.0\\
\verb+reg_alpha+ & 0.001, 0.01, 0.1, 1.0, 10.0, 100.0\\
\verb+reg_lambda+ & 0.001, 0.01, 0.1, 1.0, 10.0, 100.0\\
\verb+aft_loss_distribution_scale+ & 0.5, 0.8, 1.1, 1.4, 1.7, 2.0\\\midrule[\heavyrulewidth]
Hyperparameter & Distribution for random search\\\midrule
\verb+learning_rate+ & log uniform in $[0.001, 1.0]$\\
\verb+max_depth+ & integers in $[2, 10]$\\
\verb+min_child_weight+ & log uniform in $[0.001, 100.0]$\\
\verb+reg_alpha+ & log uniform in $[0.001, 100.0]$\\
\verb+reg_lambda+ & log uniform in $[0.001, 100.0]$\\
\verb+aft_loss_distribution_scale+ & uniform in $[0.5, 2.0]$\\\bottomrule
\end{tabular}
\end{table}

For the grid search, we try all possible ways of choosing one or two hyperparameters out of six. The generalization performance is measured in the aggregate with test accuracy.

As in Section~\ref{sec:chipseq}, we perform nested cross-validation to estimate the generalization performance of the model as well as the hyperparameter search procedure. We use 4 and 5 folds for the outer and inner cross-validation, respectively. We used data sets from Sections~\ref{sec:chipseq} and \ref{sec:simulated}.

In order to perform lots of hyperparameter search in a short amount of time, Amazon Web Services (AWS) is used to launch parallel jobs, in order to evaluate many hyperparameter search strategies. The manager EC2 instance launches hundreds of worker EC2 instances and then submits commands to execute via SSH. To ensure that all software dependencies are available to the experiment code as well as our XGBoost code, we package our code in a Docker container and host the container on Elastic Container Registry (ECR). The workers then pull the latest container image from ECR. The experiment is logged to an S3 bucket.

In all runs, the random search with 1000 trials gives the highest validation accuracy. However, high validation accuracy does not lead to high test accuracy. The grid search with one or two hyperparameters, where the number of trials is fewer than 100, yields higher test accuracy than the random search with 1000 trials. Refer to Appendix~\ref{appendix:sensitivity} to find the results for all datasets and hyperparameter search methods. When it comes to improving aggregate measure of generalization, test accuracy, it suffices to try 100 hyperparameter combinations; it does not make much difference in test accuracy to try more than 100 combinations.

\subsection{Efficient model fitting with NVIDIA GPUs}\label{sec:gpu}
XGBoost is able to utilize NVIDIA GPUs to accelerate its gradient boosting algorithm \cite{mitchell2017accelerating, ou2020outofcore}. We port the AFT loss function so that it can run on NVIDIA GPUs. To test the performance, we generate a synthetic data set with 20 million samples, by duplicating 100000 times\footnote{The duplicated rows got the same fold assignment as their originals, so that the fraction of data points belonging to each cross-validation fold remains the same.} the data \verb+simulated.model.3+ from Section~\ref{sec:simulated}. We then fit 5 XGBoost models using the 5 cross-validation folds. Each model is trained using the best hyperparameters found in Section~\ref{sec:simulated}. Table~\ref{gpu-vs-cpu} shows the timing results. In all folds, the GPU fits the model 6.1-6.7$\times$ faster than the CPU.

We used NVIDIA Quadro\textregistered~ RTX 8000 with CUDA 10.2. The GPU has 4608 cores divided
into 72 streaming multiprocessors and 48 GB GDDR6 memory.
\begin{table}
\caption{Comparing performance of CPU and GPU using the 20 million synthetic data}
\label{gpu-vs-cpu}
\centering
\begin{tabular}{rrrrr}\toprule
\multirow{2}{*}[-3pt]{Test Fold ID} & \multirow{2}{*}[-3pt]{\# boosting rounds} & \multicolumn{3}{c}{Run time (sec)} \\\cmidrule{3-5}
  & & CPU & GPU & Speedup \\\midrule
1 & 52 & 50.72 & 8.36 & 6.1$\times$\\
2 & 149 & 150.33 & 22.49 & 6.7$\times$\\
3 & 53 & 54.07 & 8.52 & 6.3$\times$\\
4 & 81 & 81.33 & 12.44 & 6.5$\times$\\
5 & 83 & 92.48 & 14.13 & 6.5$\times$\\\bottomrule
\end{tabular}
\end{table}

\section{Limitations}\label{sec:limitations}
In this section we discuss two limitations of our current approach: sensitivity to hyperparameters and prediction of survival curves.
First, even though Section~\ref{sec:sensitivity} shows that the test accuracy is not very sensitive to the choice of hyperparameters, sensitivity to hyperparameters manifests itself in other ways.
Vieira et al. \cite{xgbse2020github} present an example where two XGBoost AFT models that were fit with different values for the hyperparameter \verb+aft_loss_distribution_scale+ (and all the other hyperparameters the same) achieved nearly identical C-index metric on a validation data set but produced highly different mean survival time on the same validation set. Aggregate metrics fail to account for this phenomenon.

In addition, the XGBoost software package lacks some tools that are often used in the literature of survival analysis, such as survival curve and confidence interval computation. 
The survival curve is defined as, for each time point $t$, the proportion of the population for which the event would complete by $t$.
Although the XGBoost predict method only computes a point estimate (mean) of the survival time for each individual in the population, interested users could also compute a survival curve using the chosen AFT distribution and scale parameter.
A concern with such survival curves is that they may not be well-calibrated.
In many applications, statistical models should be well calibrated to produce accurate probabilistic predictions, which in turn let us to accurately quantify the uncertainty around the given prediction.

It is possible to mitigate the aforementioned limitations. After our code became part of the XGBoost package on August 2020, a follow-up work \cite{xgbse2020github} addressed the shortcomings mentioned above, via model stacking. The authors created a new package XGBSE that built on top of the XGBoost AFT model, where the output of the XGBoost model is used as an input to a second model that is amenable to probability calibration, such as logistic regression or nearest neighbor. A survival curve is obtained by fitting a Kaplan-Meier estimator \cite{kaplanmeier} from the output of the second model. With this approach of model stacking, the authors were able to obtain well-calibrated survival curves that are not sensitive to the choice of hyperparameters. In short, XGBSE capitalized on existing strengths of XGBoost AFT while mitigating its limitations. The authors state that they chose to build on XGBoost AFT, because of its state-of-the-art discriminating power and generalization performance as given in test metrics.

\section{Conclusion}
We implemented the Accelerated Failure Time model in XGBoost, a widely used library for gradient boosting. Using real and simulated data sets, we show that AFT in XGBoost show competitive generalization performance and run-time efficiency, both for interval-censored and right-censored data. XGBoost gives superior generalization capacity compared to linear baselines, survReg and penaltyLearning, and runs faster than the nonlinear baseline, mmit. Furthermore, AFT in XGBoost is able to take advantage of many capabilities of the ecosystem of XGBoost, such as support for NVIDIA GPUs. A future work may take advantage of integration of XGBoost into distributed computing frameworks such as Apache Spark and Dask.

Since August 2020, when our work became part of the XGBoost package, it has enabled follow-up work in open-source statistical software. Already, packages such as XGBSE and MLR3 \cite{xgbse2020github,mlr3} take advantage of the support for AFT in XGBoost. In particular, XGBSE was built on top of our work and addressed the major shortcomings (see Section 4). Usage of our software indicates real-world relevance and impact of our contribution.

\subsection*{Acknowledgements}
This work was supported by Google Summer of Code 2019.

\printbibliography

\clearpage

\appendix
\appendixpage
\pagenumbering{Roman}
\counterwithin{figure}{section}
\counterwithin{table}{section}
\counterwithin{equation}{section}

\section{Code}
The latest XGBoost package contains support for the AFT model. The source code is available at \url{https://github.com/dmlc/xgboost}.

The experiment code is available at \url{https://github.com/hcho3/XGBoostAFTPaperCode}.

\section{Tutorial}
The first step is to express the labels in the form of a range, so that every data point has two numbers associated with it, namely the lower and upper bounds for the label. For uncensored labels, use a degenerate interval of form $[a, a]$.

\subsection{Using XGBoost AFT in Python}
Collect the lower bound numbers in one array (let’s call it \verb+y_lower+) and the upper bound number in another array (call it \verb+y_upper+). Then pass the two arrays to the \verb+xgboost.DMatrix+ constructor via arguments \verb+label_lower_bound+ and \verb+label_upper_bound+:
\begin{minted}{python}
import numpy as np
import xgboost as xgb

# 4-by-2 Data matrix
X = np.array([[1, -1], [-1, 1], [0, 1], [1, 0]])

# Associate ranged labels with the data matrix.
# This example shows each kind of censored labels.
#                  uncensored    right     left  interval
y_lower = np.array([      2.0,     3.0,     0.0,     4.0])
y_upper = np.array([      2.0, +np.inf,     4.0,     5.0])
dtrain = xgb.DMatrix(X, label_lower_bound=y_lower, label_upper_bound=y_upper)
\end{minted}
Now we are ready to invoke the training API:
\begin{minted}{python}
params = {'objective': 'survival:aft', 'eval_metric': 'aft-nloglik',
          'aft_loss_distribution': 'normal',
          'aft_loss_distribution_scale': 1.20,
          'tree_method': 'hist', 'learning_rate': 0.05, 'max_depth': 2}
bst = xgb.train(params, dtrain, num_boost_round=5, evals=[(dtrain, 'train')])
\end{minted}

\subsection{Using XGBoost AFT in R}
Collect the lower bound numbers in one array (let’s call it \verb+y_lower+) and the upper bound number in another array (call it \verb+y_upper+). Then associate the two arrays with a data matrix via calls to \verb+setinfo+:
\begin{minted}{r}
library(xgboost)

# 4-by-2 Data matrix
X <- matrix(c(1., -1., -1., 1., 0., 1., 1., 0.), nrow=4, ncol=2, byrow=TRUE)
dtrain <- xgb.DMatrix(X)

# Associate ranged labels with the data matrix.
# This example shows each kind of censored labels.
#             uncensored  right  left  interval
y_lower <- c(        2.,    3.,   0.,       4.)
y_upper <- c(        2.,  +Inf,   4.,       5.)
setinfo(dtrain, 'label_lower_bound', y_lower)
setinfo(dtrain, 'label_upper_bound', y_upper)
\end{minted}
Now we are ready to invoke the training API:
\begin{minted}{r}
params <- list(objective='survival:aft', eval_metric='aft-nloglik',
               aft_loss_distribution='normal',
               aft_loss_distribution_scale=1.20,
               tree_method='hist', learning_rate=0.05, max_depth=2)
watchlist <- list(train = dtrain)
bst <- xgb.train(params, dtrain, nrounds=5, watchlist)
\end{minted}

\subsection{Note on hyperparameters}
Set \verb+objective+ hyperparameter to \verb+survival:aft+ and \verb+eval_metric+ to \verb+aft-nloglik+ in order to use the AFT model in XGBoost. The hyperparameter \verb+aft_loss_distribution+ corresponds to the distribution of $Z$ in the AFT model, and \verb+aft_loss_distribution_scale+ corresponds to the scaling factor $\sigma$. Currently, you can choose from three probability distributions for \verb+aft_loss_distribution+: \verb+normal+, \verb+logistic+, and \verb+extreme+.

\section{Full proof for Definition~\ref{def:grad-hess-aft}, the gradient and hessian of the AFT loss}\label{appendix:loss_gradient_full_proof}
We first derive the first- and second-order partial derivatives of $\ell_{\mathrm{AFT}}$ for the uncensored case, using basic rules of Calculus:
\begin{align}
\ell(y, u)
&= -\ln{\left[f_Z(s(y)) \cdot \dfrac{1}{\sigma y}\right]}\\
\frac{\partial \ell}{\partial u}
&= -\frac{\partial}{\partial u} \ln{\left[f_Z(s(y)) \cdot \dfrac{1}{\sigma y}\right]}\\
&= -\frac{1}{f_Z(s(y)) \cdot \dfrac{1}{\sigma y}}\cdot \frac{\partial}{\partial u}\left[f_Z(s(y)) \cdot \dfrac{1}{\sigma y}\right] && \text{Chain Rule}\\
&= -\frac{\sigma y}{f_Z(s(y))}\cdot \left[ f_Z'(s(y)) \cdot \frac{\partial}{\partial u} \frac{\ln{y} - u}{\sigma} \cdot \frac{1}{\sigma y}  \right] && \text{Chain Rule}\\
&= -\frac{\sigma y}{f_Z(s(y))}\cdot \left[ f_Z'(s(y)) \cdot -\frac{1}{\sigma} \cdot \frac{1}{\sigma y}  \right]\\
&= \frac{f_Z'(s(y))}{\sigma f_Z(s(y))}\\
\frac{\partial^2 \ell}{\partial u^2}
&= \frac{\partial}{\partial u}\frac{\partial \ell}{\partial u}\\
&= \frac{\partial}{\partial u}\frac{f_Z'(s(y))}{\sigma f_Z(s(y))}\\
&= \frac{\partial / \partial u [f_Z'(s(y))] \cdot \sigma f_Z(s(y)) - f_Z'(s(y)) \cdot \sigma \partial / \partial u [f_Z(s(y))] }{ \sigma^2 f_Z(s(y))^2 } &&\text{Quotient Rule}\\
&= \frac{f_Z''(s(y)) \cdot (-1/\sigma) \cdot \sigma f_Z(s(y)) - f_Z'(s(y)) \cdot \sigma f_Z'(s(y)) \cdot (-1/\sigma) }{ \sigma^2 f_Z(s(y))^2 } &&\text{Chain Rule}\\
&= -\frac{f_Z''(s(y))f_Z(s(y)) - f_Z'(s(y))^2}{\sigma^2 f_Z(s(y))^2}
\end{align}
The censored case proceeds similarly:
\begin{align}
\ell(y, u)
&= -\ln{\left[F_Z(s(\overline{y})) - F_Z(s(\underline{y}))\right]}\\
\frac{\partial \ell}{\partial u}
&= -\frac{\partial}{\partial u}\ln{\left[F_Z(s(\overline{y})) - F_Z(s(\underline{y}))\right]}\\
&= -\frac{1}{F_Z(s(\overline{y})) - F_Z(s(\underline{y}))}\cdot \frac{\partial}{\partial u} \left[F_Z(s(\overline{y})) - F_Z(s(\underline{y}))\right] && \text{Chain Rule}\\
&= -\frac{1}{F_Z(s(\overline{y})) - F_Z(s(\underline{y}))}\cdot \left[ f_Z(s(\overline{y})) \cdot -\frac{1}{\sigma} - f_Z(s(\underline{y})) \cdot -\frac{1}{\sigma} \right]&& \text{Chain Rule; $f_Z = F_Z'$}\\
&= \frac{f_Z(s(\overline{y})) - f_Z(s(\underline{y}))}{\sigma [F_Z(s(\overline{y})) - F_Z(s(\underline{y}))]}\\
\frac{\partial^2 \ell}{\partial u^2}
&= \frac{\partial}{\partial u}\frac{\partial \ell}{\partial u}\\
&= \frac{\partial}{\partial u}\frac{f_Z(s(\overline{y})) - f_Z(s(\underline{y}))}{\sigma [F_Z(s(\overline{y})) - F_Z(s(\underline{y}))]}\\
&= \frac{ \partial / \partial u [ f_Z(s(\overline{y})) - f_Z(s(\underline{y})) ] \cdot \sigma [F_Z(s(\overline{y})) - F_Z(s(\underline{y}))]}{\sigma^2 [F_Z(s(\overline{y})) - F_Z(s(\underline{y}))]^2} \notag\\
&\phantom{=}- \frac{ [f_Z(s(\overline{y})) - f_Z(s(\underline{y}))]\cdot \sigma \partial / \partial u [F_Z(s(\overline{y})) - F_Z(s(\underline{y}))]}{\sigma^2 [F_Z(s(\overline{y})) - F_Z(s(\underline{y}))]^2}&& \text{Quotient Rule}\\
&= \frac{ [ f_Z'(s(\overline{y})) - f_Z'(s(\underline{y})) ] \cdot (-1/\sigma) \cdot \sigma [F_Z(s(\overline{y})) - F_Z(s(\underline{y}))]}{\sigma^2 [F_Z(s(\overline{y})) - F_Z(s(\underline{y}))]^2} \notag\\
&\phantom{=}- \frac{ [f_Z(s(\overline{y})) - f_Z(s(\underline{y}))]\cdot \sigma [f_Z(s(\overline{y})) - f_Z(s(\underline{y}))] \cdot (-1/\sigma)}{\sigma^2 [F_Z(s(\overline{y})) - F_Z(s(\underline{y}))]^2} &&\text{Chain Rule; $f_Z = F_Z'$}\\
&= \dfrac{\Longstack[l]{ -[ F_Z(s(\overline{y})) - F_Z(s(\underline{y})) ][ f_Z'(s(\overline{y})) - f_Z'(s(\underline{y})) ] \\ \phantom{-} + [ f_Z(s(\overline{y})) - f_Z(s(\underline{y})) ]^2 } }{\sigma^2 [ F_Z(s(\overline{y})) - F_Z(s(\underline{y})) ]^2 }
\end{align}

\section{Effect of hyperparameters on model performance}\label{appendix:sensitivity}
The goal of the experiment is to quantify the impact of the hyperparameter choice on the generalization performance. For each model produced by each hyperparameter combination, we compute the test accuracy using the held-out test set.

The following tables show the validation and test accuracy of all models produced by the hyperparameter search. See Section~\ref{sec:sensitivity} for detailed instructions.

Is the AFT method very sensitive to the hyperparameter choice? If yes, then we will need to run many trials of hyperparameter search and the performance benefit of XGBoost will be negated. If not, then we will be able to run a comparatively small number of trials and we will be able to obtain an optimal model in short amount of time.

The answer to the question of the preceding paragraph is yes. For all data sets, running 1000 trials of hyperparameter search did not produce significant advantage over running only 100 trials, when measured in the test acccuracy mesure. Thus, we can simply run 100 rounds of hyperparameter search and obtain a close-to-optimal model, at least when measured in the aggregate with the test accuracy.

Unfortunately, this analysis has a gaping hole: a model being optimal in the aggregate does not mean it is well calibrated in individual predictions. Two models may have similar test accuracy but produce wildly differing prediction for some inputs. See discussion in Section~\ref{sec:limitations}.
\begin{center}
\footnotesize
\renewcommand{\arraystretch}{0.8}
\begin{longtable}{rllrrr}
\caption{Comparison of hyperparameter search methods. Both validation and test accuracy are averages over the 4 outer cross-validation folds.} \label{appendix:table:sensitivity}\\
\toprule
Rank &  Method &                    Hyperparameters selected &  \# trials &  Valid. acc. &  Test acc. \\
\toprule
\endfirsthead
\multicolumn{6}{l}{\tablename\ \thetable\ (Continued from previous page)} \\
\toprule
Rank &  Method &                    Hyperparameters selected &  \# trials &  Valid. acc. &  Test acc. \\
\toprule
\endhead
\bottomrule \multicolumn{6}{c}{(Continued on next page)} \\
\endfoot
\bottomrule
\endlastfoot
\multicolumn{6}{c}{Dataset ATAC\_JV\_adipose, normal distribution}\\
\midrule

  0 &      Grid &                          max\_depth, reg\_lambda &    54 &  0.715249 &  0.448685 \\
  1 &      Grid &                       learning\_rate, max\_depth &    36 &  0.710944 &  0.446252 \\
  2 &      Grid &                                      reg\_alpha &     6 &  0.686967 &  0.445603 \\
  3 &      Grid &                    max\_depth, min\_child\_weight &    45 &  0.708613 &  0.443379 \\
  4 &    Random &                                        All six &  1000 &  0.724057 &  0.441571 \\
  5 &      Grid &                                      max\_depth &     9 &  0.704368 &  0.440111 \\
  6 &  Baseline &                            None (all defaults) &     1 &  0.677217 &  0.435522 \\
\end{longtable}
\end{center}

\end{document}